\documentclass[sigconf]{acmart}

\AtBeginDocument{%
  }

\copyrightyear{2024}
\acmYear{2024}
\setcopyright{rightsretained}
\acmConference[]{}{}{}

\usepackage{subcaption}
\usepackage{graphicx}

\begin{document}

\title{SurfaceAI: Automated creation of cohesive road surface quality datasets based on open street-level imagery}


\author{Alexandra Kapp, Edith  Hoffmann, Esther Weigmann, Helena Mihaljević}
\affiliation{%
  \institution{Hochschule für Technik und Wirtschaft Berlin (HTW Berlin)}
  \city{Berlin}
  \country{Germany}
}

\renewcommand{\shortauthors}{Kapp et al.}

\begin{abstract}

This paper introduces SurfaceAI, a pipeline
designed to generate comprehensive georeferenced datasets on road surface type and quality from openly available street-level imagery. The motivation stems from the significant impact of road unevenness on the safety and comfort of traffic participants, especially vulnerable road users,  emphasizing the need for detailed road surface data in infrastructure modeling and analysis. 
SurfaceAI addresses this gap by leveraging crowdsourced Mapillary data to train models that predict the type and quality of road surfaces visible in street-level images, which are then aggregated to provide cohesive information on entire road segment conditions. 
\end{abstract}



\maketitle

\section{Introduction}
Road damages can have a significant impact on the comfort and safety of traffic participants, especially on vulnerable road users such as cyclists~\cite{gadsby_understanding_2022} or wheelchair users~\cite{pearlman_pedestrian_2013},  and have been identified as a relevant cause for traffic accidents~\cite{kurebwa2019study}.
Thus, comprehensive information on road surface types (e.g., paving stones) and their quality is crucial for the analysis and advancement of road infrastructure
~\cite{wage2020ride}, or for routing applications~\cite{hrnvcivr2016practical}. However, the required data sources are commonly lacking.
While OpenStreetMap (OSM)  
offers tags for surface type and quality, large gaps within the database exist as data availability depends on contributions from volunteers. 
For example, as of August 2024,  only 8,6\% of road segments in Germany are tagged with quality information.

There is a large body of research on road surface state assessment~\cite{s24175652, kim_review_2022, survey_road_monitoring_2021}, focusing primarily on  detecting damages such as cracks and potholes~\cite{zakeri2017image, ye_convolutional_2021, harikrishnan_vehicle_2017}.
Yet, road damage alone may not reflect the full range of factors that influence a traffic participant's experience. For example, the smoothness of sett (regular-shaped cobblestone) is also influenced by the flatness of the stones. Thus, a combination of road surface type and quality information is necessary to provide a comprehensive picture of the driving experience.
Satellite images have been proposed for large-scale surface type identification~\cite{zhou_mapping_2024}, but limited to distinguishing between paved and unpaved roads. Karakaya et al.~\cite{karakaya_crowdsensing_2023} used crowdsourced accelerometer data to derive surface quality information for bicycle paths and demonstrated its applicability in routing. While this approach can predict comfort, it can struggle with road type differentiation (e.g., roadway or sidewalk) in a crowdsourced setting,
and relies on sensor data which is not widely available yet.

Rateke et al.~\cite{rateke_road_2019} developed models predicting the surface type and quality from street-level imagery. The solution presented in this work incorporates image classification models of similar architecture. However, we extend their approach by utilizing a finer-grained category scheme, a more diverse dataset, and optimizations in the model architecture. Moreover, to the best of our knowledge, no existing works have presented a pipeline enabling the prediction of road surface type and quality on a large scale for comprehensive road networks across arbitrary municipalities using available data. 
We aim to fill this gap by presenting  \textit{SurfaceAI}, a process pipeline and a software that uses openly available street-level imagery to generate georeferenced datasets pertaining to surface type and quality across arbitrary areas.

\begin{figure*}[tb]
    \centering
    \includegraphics[width=0.9\textwidth]{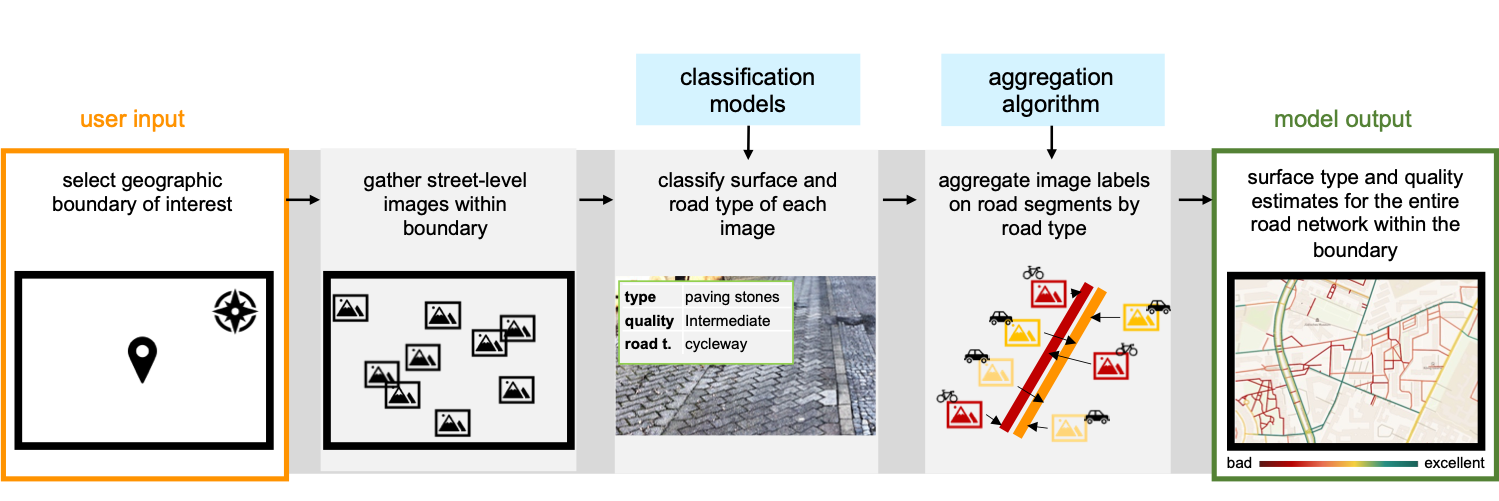}
    \caption{Schematic of the proposed pipeline from user input (geographic boundary) to model output (classified road network). Note that, for clarity, the diagram displays the
     aggregation algorithm for quality only. }
    \label{fig:pipeline}
\end{figure*}

\section{Proposed approach}

Street-level images are collected based on a user-defined geographic boundary of interest. These images are then classified by road type, surface type, and surface quality, and aggregated at the road segment level to provide estimates for the entire road network within the defined boundary (see Figure~\ref{fig:pipeline}). The individual steps of the pipeline are elaborated below.

We implement the pipeline as open-source code in Python, using PyTorch~\cite{NEURIPS2019_9015} for neural network models and PostGIS~\cite{postgis2018postgis} for efficient spatial operations\footnote{Github Repository: \url{https://github.com/SurfaceAI/road_network_classification}}.

\subsection{Street-level imagery}
Our approach leverages   \textit{Mapillary}~\cite{noauthor_mapillary_2024}, a crowdsourcing platform launched in 2013 that provides openly available street-level images. 
As of January 2024, the dataset contains about 170 million images in Germany, with over 50\% captured within the past three years.
Its smartphone app allows contributors to easily capture georeferenced image sequences during their trips by car, bicycle, or on foot, resulting in a diverse collection that includes roadways, cycleways, and footways. Additionally, Mapillary offers an API~\cite{noauthor_mapillary_api_nodate} for programmatic access, which we integrate into our software stack.

The accessibility and widespread use of Mapillary, along with the heterogeneity of the data regarding  contributors, regions, devices, camera angles, and transportation modes, make it a suitable choice for training  robust image classification models. Imagery from Mapillary, along with its metadata and data extracted using computer vision methods, has been employed for various use cases, including dataset creation for urban analytics covering a variety of traffic participants views~\cite{hou_global_2024, neuhold_mapillary_2017}, generation of a cycleway network~\cite{ding_towards_2021}, and spatiotemporal analysis of city dynamics, e.g., in the context of walkability~\cite{wang_investigating_2024}.

Other accessible data sources could be easily integrated into our pipeline, supplementing or even replacing Mapillary as the image database.   
Crowdsourcing platforms such as KartaView~\cite{noauthor_kartaview_nodate} and Panoramax~\cite{noauthor_panoramax_nodate}, the latter currently used mainly in France, provide similar types of images, but have a substantially lower coverage in Germany. The commercial image platform Google Street View~\cite{noauthor_google_nodate} offers panorama imagesgh quality due to the prevalence of standardized recording by the platform itself with professional equipment, but the data is not freely available. In addition, Google halted image publication in Germany for several years,  only resuming in mid-2023. 
Moreover, unlike Mapillary, contribution by volunteers is possible only with advanced technology for recording panorama images.

\subsection{Image classification models}

We develop a supervised deep learning-based model that predicts, from a Mapillary image, (1) the \textit{surface type} (e.g., `asphalt') and (2) the \textit{surface quality} (e.g., `intermediate') which reflects the physical usability of a road segment for wheeled vehicles, particularly regarding its regularity or flatness.
Specifically, we fine-tune EfficientNetV2-S~\cite{tan2021efficientnetv2}, pre-trained on ImageNet, using the recently published dataset StreetSurfaceVis~\cite{kapp_2024_11449977, kapp2024streetsurfacevisdatasetcrowdsourcedstreetlevel} comprising 9,122 Mapillary images manually annotated by surface type and quality. 

The labeling scheme closely aligns with that of OSM, with surface type values consisting of \textit{asphalt}, \textit{concrete}, \textit{paving stones}, \textit{sett}, and \textit{unpaved}, while quality values range from \textit{excellent} and \textit{good} to  \textit{intermediate}, \textit{bad} and  \textit{very bad}. For more details on the annotation process, refer to~\cite{kapp2024streetsurfacevisdatasetcrowdsourcedstreetlevel}. We train a classification model to predict the surface type, and one regression model per surface type predicting the surface quality. 

According to evaluations from \cite{kapp2024streetsurfacevisdatasetcrowdsourcedstreetlevel}, for an 80:20 train-validation split and a test set comprising 776 images from geographically distinct cities, the type classification model performs well, achieving an accuracy (loss) of 0.96 (0.13) on the training data, 0.94 (0.19) on the validation data, and 0.91 on the test data. The F1 scores for individual type classes in the test data are all equal to or exceed 0.9, except for the `concrete' surface type, resulting in a weighted average F1 score of 0.84.  These results indicate that the model generalizes effectively across different locations. For the quality regression models, deviations from the true values are normally distributed and centered around 0, suggesting no systematic bias in the predictions. Regarding quality predictions, the overall Spearman correlation coefficient of 0.72 (ranging from .42 to .65 for individual type classes), an accuracy of 0.63, and a 1-off accuracy (considering neighboring classes as correct classifications) of almost 1.0 reflect a strong positive relationship between the predicted and true quality rankings. While the model effectively captures the relative ordering of surface quality,  some variability remains, similar to human assessments, as quality classes are rather fluid~\cite{kapp2024streetsurfacevisdatasetcrowdsourcedstreetlevel}. 

Since roads, especially in urban settings, often consist of various sections such as roadways, cycleways, sidewalks, or other areas like parking lots or green stripes, we train an additional model  to resolve this ambiguity. Specifically, we fine-tune a classifier, with an architecture similar to the surface type model, to distinguish between the following \textit{road type} classes: \textit{roadway}, \textit{bike lane}, \textit{cycleway}, \textit{sidewalk}, \textit{path}, and \textit{no road or no single focus area}. The last category includes images that either lack a clear focus on any part of the road or mainly depict non-road elements, such as buildings or cars.  

Based on 7,324 images with a train-validation split of 80:20, an initial model achieves an accuracy (loss) of 0.99 (0.03) for the training data and 0.88 (0.45) on the validation dataset. A weighted F1 score of 0.88 is reached on the validation data, with class-specific F1 scores of 0.95 for \textit{roadway}, 0.87 for \textit{bike lane}, 0.76 for \textit{cycleway}, 0.59 for \textit{sidewalk}, 0.91 for \textit{path}, and 0.82 for \textit{no road or no single focus area} on the validation dataset.

\subsection{Aggregation algorithm}

To obtain a cohesive dataset for the entire road network within a geographic boundary, an algorithm aggregates classifications of individual images. 
We use the road segments provided in OSM as predefined aggregation units without further subdivisions, ensuring better compatibility with the utilized geographic database. Additionally, relying solely on class predictions can make it difficult to determine whether variations within a single road segment reflect actual differences in surface type or quality, which would justify a further splitting of a segment, or if they result from noise through factors such as imprecise geotags, incorrect classifications, or inconsistent surfaces. 

\begin{figure}[tb]
  \centering
  \begin{subfigure}{0.22\textwidth}
    \centering
    \includegraphics[width=\textwidth]{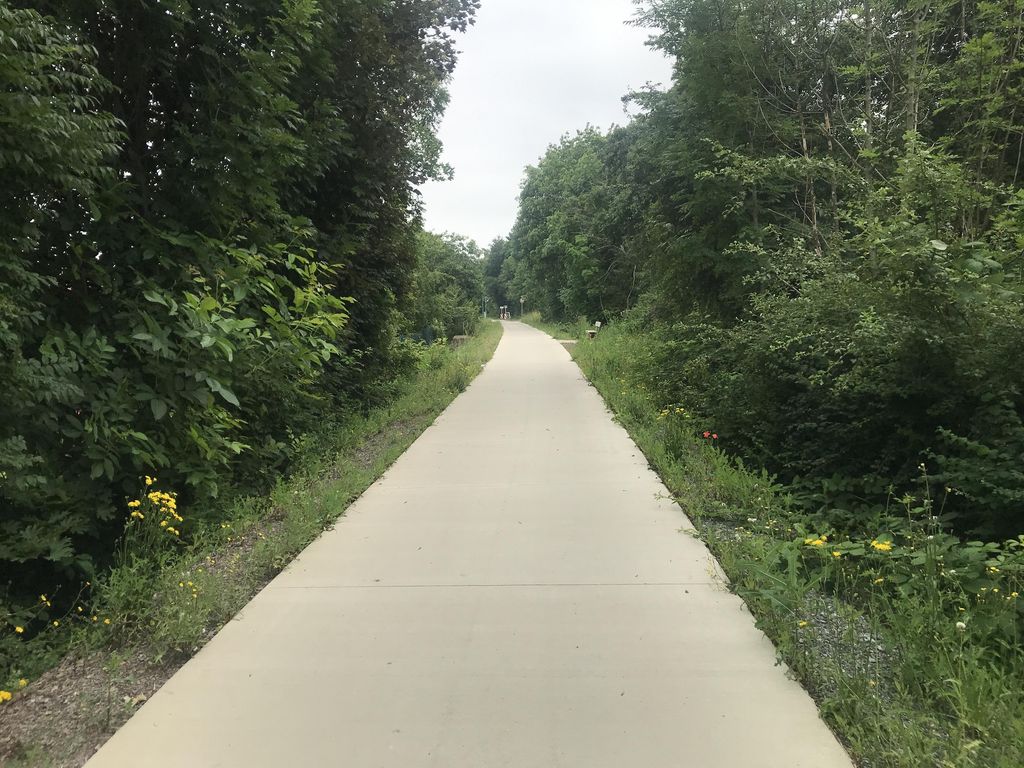}
    \caption{\footnotesize{road type unclear from single image} \newline \tiny{surveyor99|123502363306902}}
    \end{subfigure}
  \begin{subfigure}{0.22\textwidth}
    \centering
    \includegraphics[width=\textwidth]{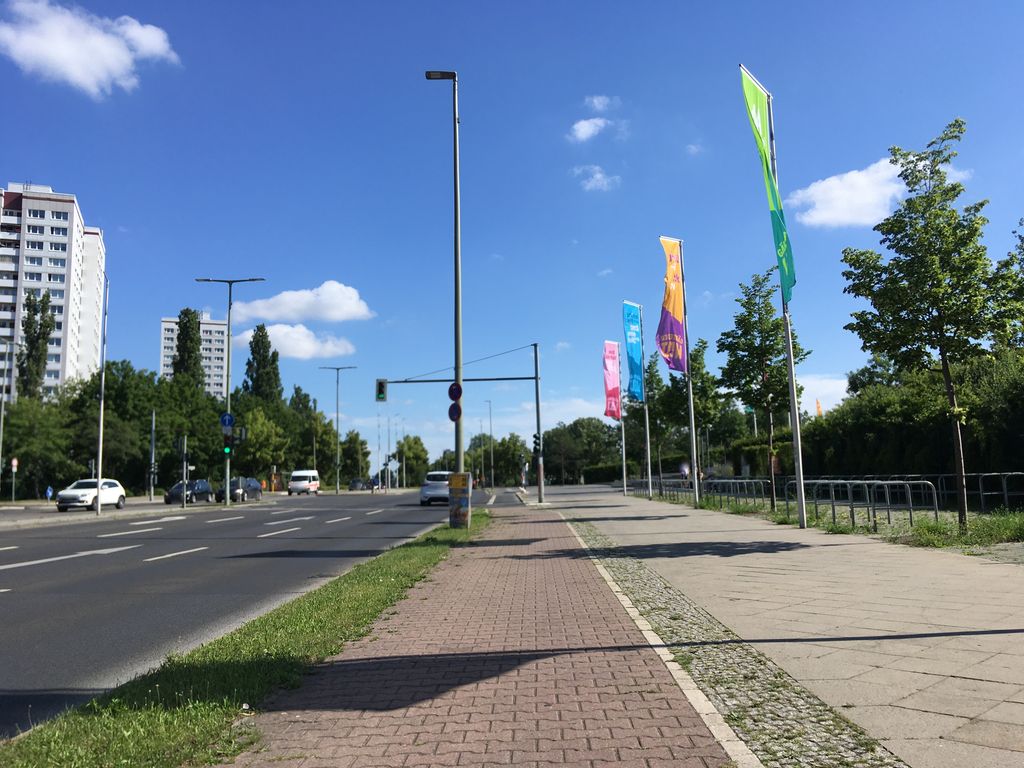}
    \caption{\footnotesize{road with three road types: roadway, cycleway, and sidewalk} \newline \tiny{carlheinz|160748322585932}}
  \end{subfigure}
  \caption{Examples of clear and ambiguous road types. Images from Mapillary; contributor names and image IDs are indicated for each image.}
  \label{fig:exampleImg}
\end{figure}

Images within the region of interest are assigned to a road segment based on their geo-coordinates. If an image is near multiple road segments, such as a roadway and a cycleway (cf. image (b)  in Figure \ref{fig:exampleImg}), we use the predicted road type class for the respective images to resolve any ambiguity. Using all images successfully assigned to a road segment present in the underlying OSM network, the surface type is determined by majority vote, while the surface quality of the segment is calculated as an average.

Road segments may span multiple hundred meters, and certain parts may be overrepresented, such as when a certain driveway is highly frequented by one contributor. To prevent single parts from disproportionately influencing the overall road segment rating, we adjust the presented base idea by first computing aggregated values for 20-meter subsegments and then calculating unweighted aggregates of the values corresponding to these subsegments as the predicted value for the entire road segment. To remove unreliable results, a subsegment is required to have at least three agreeing type classifications, and a road segment to have classifications for at least half their subsegments, otherwise the value is considered missing. 

It should be noted that, while OSM is a viable choice due to its high world-wide coverage and easy accessibility, other geographic databases can be substituted in the aggregation algorithm, particularly if they offer a more reliable or finer-grained network for the area of interest.

\section{Evaluation in real-world scenarios}
\label{sec:real_world}

The pipeline development was informed by communication with German municipalities that could benefit from the enriched road network data. Specifically, we assessed the surface state on a provided \textbf{target cycleway network of Berlin}.
Information on whether a cycleway is present on a road segment or if cyclists should use the roadway was provided for 80\% of the segments. For the remaining 20\%, this information was inferred based on road type classifications.
Based on a manually labeled sample of this dataset, we obtained an accuracy of 0.91 and an average F1 score of 0.84 for surface type, along with a Spearman correlation of 0.53 and an accuracy of 0.62 for surface quality. These results suggest strong generalizability of the type model and a robust performance of the aggregation procedure. 
A more sensible aggregation algorithm in future work for surface quality could likely bear potential for performance improvements.
Additionally, 20\% of the road segments of the entire cycling network lacked classification values due to missing images. Thus, even in a large city like Berlin, where Mapillary has comparably very good coverage, certain parts of the infrastructure remain underrepresented.

Municipalities and other interested parties can address these gaps to fully utilize the pipeline. 
One such example is a  \textbf{rural municipality} in Germany who captured all their roads using a car-based setup and uploaded the images to Mapillary. As the images were captured consistently and in a high-quality homogeneous setup, issues related to image quality, recency, completeness, and road type classification were significantly reduced. 
However, while asphalt and unpaved roads were well distinguished, the type classification model struggled with paving stone surfaces. (Concrete and sett were excluded from this evaluation due to negligible occurrence in the dataset). This difficulty was likely caused by the different characteristics of rural roads compared to the predominantly urban settings in the training data, revealing limitations in transferability.
Specifically, based on a manually labeled sample of 214 road segments, we obtained an accuracy of 0.76, with F1 scores of  0.94 for asphalt,  0.74 for unpaved, and 0.58 for paving stones. A Spearman correlation  of 0.52 for surface quality and an  accuracy of 0.66 align well with the evaluation of the Berlin use case. 

\section{Limitation and discussion}

This paper proposes a pipeline that uses crowdsourced street-level imagery to classify road network surface type and quality.
While initial evaluations in real-world scenarios demonstrated the feasibility of this approach, more extensive testing is required to assess the overall performance. 
The described pipeline constitutes an initial implementation, but several limitations and areas for improvement remain. 

Firstly, the underlying image database, 
Mapillary, does not guarantee  good coverage across all regions, 
with varying coverage of individual municipalities depending on contributor activity. Additionally, image quality may vary, the recency of images might not be provided, and images taken from a car perspective might be over-represented, potentially introducing bias toward certain road types. However, the infrastructure offered by the Mapillary app provides an easily accessible option for additional data collection, which can be utilized by communities, as demonstrated in our second example in Section \ref{sec:real_world}. 

Secondly, the current implementation relies on pre-defined road segments. However, road segments can exhibit non-uniform surface characteristics, and future iterations should include methods to subdivide segments when surface changes are detected in a robust and reliable manner.

Thirdly, the current aggregation mechanism for surface quality simply averages all values, without considering how different qualities may influence the overall rating. For example, if a subsegment is rated as `bad', the entire road segment might need to obtain a low rating. As our initial evaluation for the Berlin cycleway network in Berlin in Section~\ref{sec:real_world} suggests, further research into more meaningful aggregation may be beneficial. 

Moreover, further research into road type classification could enhance accuracy and make it independent of underlying road network metadata. While current road scene segmentation models~\cite{Porzi_2019_CVPR} fail to provide robust classifications on our training data, advancements in this field may offer superior results compared to image classification, as they provide more granular information. Alternatively, including additional training data sources, such as the newly released \textit{Global Streetscapes}~\cite{hou_global_2024} street-view image dataset containing road type information, may improve model performance.

The generalizability of the classification models presents an additional challenge. While the training dataset StreetSurfaceVis is mindfully created to provide a heterogeneous set of images, the model will most likely not classify every potential road type equally well. A process should be implemented to extend the training data with cases the model struggles with, including, e.g., road types and surfaces prevalent in other countries. 

Generally, transferability to other regions may pose an issue due to geographical limitation of the training data, the coverage of Mapillary images, and the coverage of OSM road network data, including metadata on road types. Additionally, our approach may encounter difficulties in regions experiencing rapid changes in their road network, as the images need to be temporally aligned.

\bibliographystyle{ACM-Reference-Format}
\bibliography{references}

\end{document}